\documentclass[pdflatex,sn-mathphys]{sn-jnl}% Math and Physical Sciences Reference Style
\jyear{2021}%

\theoremstyle{thmstyleone}%
\theoremstyle{thmstyletwo}%

\theoremstyle{thmstylethree}%

\usepackage{derivative}
\usepackage{tabularx}

\usepackage{xr}
\makeatletter

\newcommand*{\addFileDependency}[1]{% argument=file name and extension
\typeout{(#1)}% latexmk will find this if $recorder=0
\@addtofilelist{#1}
\IfFileExists{#1}{}{\typeout{No file #1.}}
}\makeatother

\usepackage{subcaption}
\usepackage[section]{placeins}
\usepackage{multirow}
\usepackage{CJKutf8}
\usepackage{graphicx} %Loading the package
\graphicspath{{figures/}} %Setting the graphicspat

\usepackage{float}
\usepackage{etoolbox}
\makeatletter
\patchcmd{\ps@headings}
{\hbox to \hsize{\hfill Springer Nature 2021 \LaTeX\ template\hfill}}
{\hbox to \hsize{}}
{}
{}
\patchcmd{\ps@titlepage}
{\hbox to \hsize{\hfill Springer Nature 2021 \LaTeX\ template\hfill}}
{\hbox to \hsize{}}
{}
{}
\makeatother
\makeatletter
\patchcmd{\ps@headings}
{\hbox to \hsize{\hfill Springer Nature 2021 \LaTeX\ template\hfill}}
{\hbox to \hsize{}}
{}
{}
\patchcmd{\ps@headings}
{\hbox to \hsize{\hfill Springer Nature 2021 \LaTeX\ template\hfill}}
{\hbox to \hsize{}}
{}
{}
\patchcmd{\ps@titlepage}
{\hbox to \hsize{\hfill Springer Nature 2021 \LaTeX\ template\hfill}}
{\hbox to \hsize{}}
{}
{}
\makeatother

\begin{document}

\begin{CJK*}{UTF8}{gbsn}

\title[FuXi-ENS]{FuXi-ENS: A machine learning model for medium-range ensemble weather forecasting}

\author[1]{\fnm{Xiaohui} \sur{Zhong}}\email{x7zhong@gmail.com}
\equalcont{These authors contributed equally to this work.}
\author[1,2]{\fnm{Lei} \sur{Chen}}\email{cltpys@163.com}
\equalcont{These authors contributed equally to this work.}

\author*[1,2]{\fnm{Hao} \sur{Li}}\email{lihao$\_$lh@fudan.edu.cn}
\equalcont{These authors contributed equally to this work.}

\author[1]{\fnm{Jun} \sur{Liu}}\email{liujun$\_$090003@163.com}

\author[2]{\fnm{Xu} \sur{Fan}}\email{fanxu@sais.com.cn}

\author[3]{\fnm{Jie} \sur{Feng}}\email{fengjiefj@fudan.edu.cn}

\author[4]{\fnm{Kan} \sur{Dai}}\email{daikan@cma.gov.cn}

\author[5]{\fnm{Jing-Jia} \sur{Luo}}\email{jjluo@nuist.edu.cn}

\author[6]{\fnm{Jie} \sur{Wu}}\email{wujie@cma.gov.cn}

% \author[1,2]{\fnm{Yuan} \sur{Qi}}\email{qiyuan@fudan.edu.cn}

\author*[6]{\fnm{Bo} \sur{Lu}}\email{bolu@cma.gov.cn}

%\author*[2,1]{\fnm{Yuan} %\sur{Qi}}\email{qiyuan@fudan.edu.cn}

\affil[1]{\orgdiv{Artificial Intelligence Innovation and Incubation Institute}, \orgname{Fudan University}, \orgaddress{\city{Shanghai}, \postcode{200433}, \country{China}}}

\affil[2]{\orgname{Shanghai Academy of Artificial Intelligence for Science}, \orgaddress{\city{Shanghai}, \postcode{200232}, \country{China}}}

\affil[3]{\orgdiv{Department of Atmospheric and Oceanic Sciences and Institute of Atmospheric Sciences}, \orgname{Fudan University}, \orgaddress{\city{Shanghai}, \postcode{200433}, \country{China}}}

\affil[4]{\orgdiv{National Meteorological Information Center}, \orgname{China Meteorological Administration}, \orgaddress{\city{Beijing}, \postcode{100081}, \country{China}}}

\affil[5]{\orgdiv{Institute for Climate and Application Research (ICAR)/CIC-FEMD/KLME/ILCEC}, \orgname{Nanjing University of Information Science and Technology}, \orgaddress{\city{Nanjing}, \postcode{210044}, \country{China}}}

\affil[6]{\orgdiv{China Meteorological Administration Key Laboratory for Climate Prediction Studies}, \orgname{National Climate Center}, \orgaddress{\city{Beijing}, \postcode{100081}, \country{China}}}


\abstract{
Ensemble forecasting is crucial for improving weather predictions, especially for forecasts of extreme events. Constructing an ensemble prediction system (EPS) based on conventional numerical weather prediction (NWP) models is highly computationally expensive. Machine learning (ML) models have emerged as valuable tools for deterministic weather forecasts, providing forecasts with significantly reduced computational requirements and even surpassing the forecast performance of traditional NWP models. However, challenges arise when applying ML models to ensemble forecasting. Recent ML models, such as GenCast and SEEDS model, rely on the ERA5 Ensemble of Data Assimilations (EDA) or operational NWP ensemble members for forecast generation. Their spatial resolution is also considered too coarse for many applications. To overcome these limitations, we introduce FuXi-ENS, an advanced ML model designed to deliver 6-hourly global ensemble weather forecasts up to 15 days. This model runs at a significantly increased spatial resolution of 0.25\textdegree, incorporating 5 atmospheric variables at 13 pressure levels, along with 13 surface variables. By leveraging the inherent probabilistic nature of Variational AutoEncoder (VAE), FuXi-ENS optimizes a loss function that combines the continuous ranked probability score (CRPS) and the Kullback–Leibler (KL) divergence between the predicted and target distribution, facilitating the incorporation of flow-dependent perturbations in both initial conditions and forecast. This innovative approach makes FuXi-ENS an advancement over the traditional ones that use L1 loss combined with the KL loss in standard VAE models for ensemble weather forecasting. Evaluation results demonstrate that FuXi-ENS outperforms ensemble forecasts from the European Centre for Medium-Range Weather Forecasts (ECMWF), a world leading NWP model, in the CRPS of 98.1\% of 360 variable and forecast lead time combinations. This achievement underscores the potential of the FuXi-ENS model to enhance ensemble weather forecasts, offering a promising direction for further development in this field.
}

%A unique feature of FuXi-S2S is its perturbation module, which introduces flow-dependent perturbations in hidden features, enhancing forecast uncertainty estimation and subseasonal forecast skill. 


\maketitle

\section{Introduction}
Weather forecasts are inherently uncertain due to the chaotic nature of the highly nonlinear atmosphere \cite{Lorenz1963}, necessitating the incorporation of uncertainty assessments for an improved forecast \cite{national2006completing}. For various applications, forecasts that include uncertainty estimates are usually more valuable \cite{scher2018predicting,calvo2024real}. This is particularly true in sectors sensitive to weather and climate conditions for risk assessment \cite{Palmer2002,GOODARZI2019}, in renewable energy forecasting to ensure reliable and cost-effective operations of power system \cite{SPERATI2016,wang2017deep,wu2018probabilistic}, and in the aviation industry for enhanced safety and efficiency \cite{zhang2017probabilistic}. The primary sources of uncertainty arise from two main factors: imperfections in forecast models that fail to accurately simulate all atmospheric processes, and inaccuracies in initial conditions due to observational data capturing only a limited subset of atmospheric information and deficiencies in data assimilation systems. 

To date, ensemble forecasting has proven to be the most promising method for estimating forecast uncertainty, primarily through multiple runs of numerical weather prediction (NWP) models. Each run introduces slight variations in initial conditions and model physics \cite{gneiting2005weather,leutbecher2008ensemble}, effectively addressing the major sources of forecast uncertainty. 
For instance, the European Centre for Medium-Range Weather Forecasts (ECMWF) ensemble prediction system (EPS), a globally recognized leader in operational medium-range EPS, employs the singular vector (SV) approach \cite{Molteni1996,Buizza1995,barkmeijer19993d} and the stochastically perturbed parametrization tendency (SPPT) scheme \cite{buizza1999stochastic,palmer2009stochastic} to introduce perturbations to initial conditions and physical tendencies.
The SV method identifies the most sensitive directions within the initial conditions, where even minor modifications can result in significantly different forecasts, proving particularly invaluable for medium-range weather forecasting.

Despite the efficacy of convectional EPS, ensemble forecasts are computationally expensive, often necessitating a compromise in model spatial resolution and/or the number of ensemble members. Historically, the ECMWF has performed ensemble forecasts at a lower resolution than its deterministic counterpart. Recent advancements in high-performance computing have enabled high-resolution ensemble forecasting at a spatial resolution of 9 km, matching that of the high-resolution forecast (HRES) \footnote{On June 27th, 2023, high-performance computing (HPC) upgrades facilitated high-resolution ensemble at a spatial resolution of 9 km, equivalent to the HRES.}.
Moreover, the ensemble approach is crucial for data assimilation, effectively enhancing forecast reliability by capturing uncertainties in background through the spread of ensemble members. However, the substantial computational costs limit the number of ensemble members that can be operationally conducted, constraining the ensembles' ability to fully represent the probability distribution of possible weather scenarios. Despite these challenges, increasing the ensemble size has been shown to improve forecast performance \cite{buizza1999stochastic,Buizza2019}. Nevertheless, computational constraints typically limit global ensemble forecasts to between 14 and 51 members across major forecasting centers \cite{leutbecher2019ensemble,lang2021more,GEFS2022}. Therefore, developing more computationally efficient methods for producing ensemble weather forecasts remains a vital area of research.
%Introduce how ensemble is done using NWP models

Recent advancements in machine learning (ML) have significantly enhanced the computational efficiency and accuracy of global weather forecasting. ML-based weather forecasting models now provide promising alternatives to traditional NWP models \cite{pathak2022fourcastnet,bi2022panguweather,lam2022graphcast,chen2023fuxi,chen2023fengwu,bouallegue2024aifs}, often matching or even surpassing the forecast skill of the high-resolution forecast (HRES) produced by ECMWF \cite{de2023machine}. Initially, ML applications in this field focused on deterministic forecasts with an emphasis on minimizing errors such as the mean absolute error (MAE) or root mean square error (RMSE), which limited the assessment of forecast uncertainties. Subsequent studies have ventured into the more complex and challenging domain of ensemble forecasting. 
Chen et al. \cite{chen2023fuxi} employed random Perlin noise to perturb initial conditions, which are independent of the background flow. This approach led to a reduction in ensemble spread from the 9th day of forecast onward. Price et al. \cite{price2023gencast} developed GenCast, a diffusion model \cite{sohl2015deep,karras2022elucidating} capable of generating ensemble forecasts by sampling from a joint probability distribution of potential weather scenarios across space and time. GenCast can produce ensemble forecasts for up to 15 days at a spatial resolution of 0.25\textdegree and a temporal resolution of 12 hours, showing superior performance compared to ECMWF ensemble forecasts. Unlike previous ML models that relied only on the ECMWF ERA5 reanalysis dataset \cite{hersbach2020era5}, GenCast also incorporates the ERA5 Ensemble of Data Assimilations (EDA), which includes 9 perturbed members and 1 control member at a spatial resolution of 0.5625\textdegree to account for uncertainties in initial conditions. However, GenCast's dependence on the ERA5 EDA poses limitations for operational forecasting. Additionally, Li et al. \cite{SEEDS2024} proposed Scalable Ensemble Envelope Diffusion Sampler (SEEDS), which generates large ensembles using two members from the Global Ensemble Forecast System (GEFS) version 12 \cite{GEFS2022}, the operational ensemble NWP system of the United Sates. SEEDS consists of two models: the generative ensemble emulation (SEEDS-GEE) model, which emulates the distribution of GEFS, and the generative postprocessing (SEEDS-GPP) model, which corrects biases in the GEFS by integrating distributions from GEFS and ERA5 EDA. However, SEEDS requires two operational NWP ensemble member for forecast generation and includes only 8 variables at a 2\textdegree resolution. Brenowitz et al. \cite{brenowitz2024practical} constructed ensembles using lagged ensemble forecasts (LEF) \cite{hoffman1983lagged}. The LEF method, while simple to implement and free from model training, is limited by the number of their ensemble members that can be generated. Inspired by probabilistic nature of Variational AutoEncoder (VAE) \cite{doersch2016tutorial,zhao2017learning}, SwinVRNN \cite{hu2022swinvrnn} and FuXi-S2S \cite{chen2023fuxis2s} focus on medium-range and sub-seasonal ensemble forecasting, respectively. These models combine reconstruction loss and the Kullback–Leibler (KL) divergence in their loss functions, introducing perturbations into the latent space to enhance forecast reliability. Despite its pioneering performance in medium-range ensemble weather forecasting, SwinVRNN produces forecasts at a 5.625\textdegree resolution, which is too coarse for many applications.
While ML models have made significant efforts in ensemble weather forecasting, the optimal approach for ML-based ensemble forecasting remains an open question. Additionally, it is unclear when ML-based ensemble weather forecasting models will consistently outperform traditional ECMWF ensembles, particularly in accurately predicting the uncertainty and the likelihood of high-impact weather events at fine spatial resolutions.
%summarize the metholodology used in ML models for ensemble forecasting.

%To summarize the methods applied before, they can be categorized into 3 categories: 1) random noise perturbations on initial conditions; 2) applying diffusion model; 3) use generative model to emulate the physics-based ensemble forecasts. 

In this paper, we introduce FuXi-ENS, a ML-based medium-range ensemble weather forecasting model that outperforms the ECMWF ensemble at a fine spatial resolution of 0.25\textdegree. FuXi-ENS generates 6-hourly forecasts up to 15 days, encompassing 5 atmospheric variables at 13 pressure levels and 13 surface variables. This model uniquely combines the continuous ranked probability score (CRPS) and KL loss, a method superior to the traditional usage of L1 loss paired with the KL loss in classical VAE model. Specifically, CRPS proves more effective than L1 loss for ensemble weather forecasting. Furthermore, FuXi-ENS introduces perturbations at both the initial conditions and each forecast step. These perturbations mirror the ensemble generation process in NWP models by accounting for errors in both initial conditions and physical tendencies. FuXi-ENS is developed using 17 years of 6-hourly ECMWF ERA5 reanalysis data at a spatial resolution of $0.25^{\circ}$. It employs analysis data and a deterministic model for initialization, complemented by a perturbation module that introduces perturbations at initialization step and every subsequent forecast steps. This distinguishes FuXi-ENS from models like SEEDS and GenCast, which rely on either EDA or operational ensemble forecast members.

FuXi-ENS demonstrates remarkable performance, outperforming the ECMWF ensemble across multiple metrics, including the accuracy of the ensemble mean forecast, ensemble verification metrics, and the prediction of extreme weather events.
Extreme weather events have significantly impacted, and will continue to impact society and the economy \cite{mirza2003climate,jahn2015economics,frame2020climate}. Recent decades have witnessed a notable increase in the frequency of such events globally, including heat wave and extreme rainfall, due to climate change \cite{national2016attribution,newman2023global,Aigner2023}.
Accurately predicting these events remains a significant and long-standing challenge in weather forecasting. Among these, tropical cyclones (TCs), as one of the most destructive extreme events, pose severe threats to human lives and property \cite{mousavi2011global,lang2012impact}. Enhanced accuracy in forecasting TC track can reduce unnecessary warnings and evacuations, thereby improving emergency preparedness and management \cite{hamill2011global,conroy2023track}. Additionally, research suggests that anthropogenic climate warming and rising sea surface temperatures (SSTs) are likely to intensify TCs and increase their associated rainfall \cite{pielke2005hurricanes,kossin2013trend,patricola2018anthropogenic,Knutson2019}.
Therefore, this studies conduct a comprehensive evaluation of probabilistic TC track forecasts using the FuXi-ENS and ECMWF ensemble models, finding FuXi-ENS's performance comparable to that of ECMWF ensemble.
Moreover, the computational efficiency of the FuXi-ENS model is remarkable. It completes a 15-day forecast with a 6-hourly temporal resolution in approximately 10 seconds per member on an Nvidia A100 graphics processing unit (GPU), which is negligible compared to conventional NWP models. The model's high spatial resolution and rapid computation speed enable a broad range of practical applications, capitalizing on the fast and accurate ensemble forecasts provided by FuXi-ENS.

\section{Results}

This study presents a comprehensive evaluation of the 48-member FuXi-ENS forecasts in forecasting weather during the testing period of one year in 2018. Each of the 8 A100 GPUs contributes to the ensemble by generating 6 members, collectively producing a total of 48 members. We compare the performance of FuXi-ENS with that of the 51-member ECMWF ensemble forecasts, utilizing a variety of metrics. These metrics encompass deterministic skill scores for ensemble mean forecasts, probabilistic forecast metrics derived from all ensemble members, and particular evaluations for extreme weather events, with a focus on probabilistic forecasts of TC tracks and the record-breaking 2018 heat wave in Northeast Asia. 
%Furthermore, the study investigates whether the forecast uncertainty can be represented by the magnitude of ensemble spread generated by the FuXi-ENS model.

\subsection{Deterministic metrics}

Deterministic metrics such as $\textrm{RMSE}$ and anomaly correlation coefficient ($\textrm{ACC}$) are crucial for evaluating ensemble mean weather forecasts \cite{murphy1989skill,hamill2013noaa}. $\textrm{RMSE}$ quantifies the average magnitude of forecast errors, providing a thorough assessment of accuracy, whereas $\textrm{ACC}$ evaluates the correlation between predicted and observed anomalies, reflecting the model's ability to capture large-scale synoptic patterns. 

\begin{figure}[h]
    \centering
    \includegraphics[width=\linewidth]{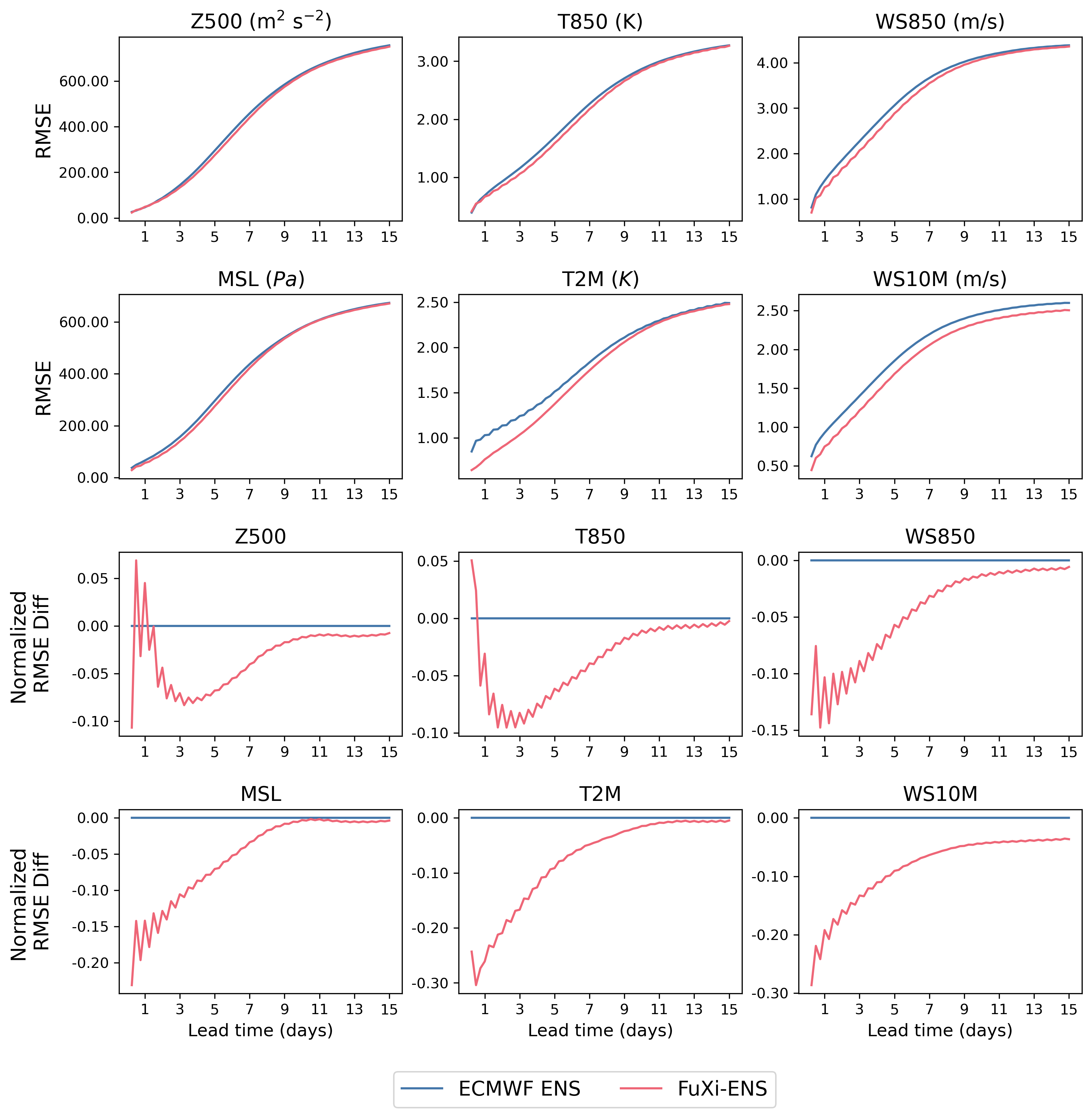}
    \caption{Comparison of the globally-averaged latitude-weighted $\textrm{RMSE}$ (first and second rows) as well as normalized $\textrm{RMSE}$ (third and fourth rows) differences of ensemble mean forecasts from the ECMWF ensemble (blue lines) and FuXi-ENS (red lines) for 3 upper-air variables, including $\textrm{Z500}$, $\textrm{T850}$, and $\textrm{WS850}$, and 3 surface variables, such as $\textrm{MSL}$, $\textrm{T2M}$, and $\textrm{WS10M}$, in 15-day forecasts using testing data from 2018. The normalized differences are calculated using the ECMWF ensemble mean as the reference.}
    \label{RMSE_plot}    
\end{figure}

This subsection presents a comparative analysis of ensemble mean forecasts from the FuXi-ENS and ECMWF ensemble using these metrics. Figure \ref{RMSE_plot} shows a time series of the globally-averaged, latitude-weighted $\textrm{RMSE}$ for both models in $\textrm{RMSE}$ at lead times of 0-15 days for 6 variables: geopotential at 500 hPa ($\textrm{Z500}$), temperature at 850 hPa ($\textrm{T850}$), wind speed at 850 hPa ($\textrm{WS850}$), mean sea level pressure ($\textrm{MSL}$), 2-meter temperature ($\textrm{T2M}$), and wind speed at 10 meter ($\textrm{WS10M}$). Due to subtle differences between the FuXi-ENS and ECMWF ensemble, normalized differences in $\textrm{RMSE}$ are also provided for enhanced clarity. Additionally, Supplementary Figure 1 illustrates the absolute $\textrm{ACC}$ values and their normalized differences as a function of forecast lead times for the same 6 variables in Figure \ref{RMSE_plot}. Many other atmospheric variables at different pressure levels are not included in the skill comparisons due to the unavailability of data in the ECMWF archive. Furthermore, acquiring all 51 members of the ECMWF ensemble from the ECMWF archive is challenging due to the substantial data volumes involved, which can not be easily downloaded. Normalized differences in $\textrm{RMSE}$ and $\textrm{ACC}$ are calculated using the ECMWF ensemble mean as the reference. 
FuXi-ENS forecasts demonstrate persistently lower $\textrm{RMSE}$ and higher $\textrm{ACC}$ skill than the ECMWF ensemble mean for nearly all the evaluated variables and forecast lead times, except for $\textrm{Z500}$ and $\textrm{T850}$, where the ensemble mean of FuXi-ENS is less accurate than that of ECMWF ensemble at 2 out of 60 forecast lead times. 

Supplementary Figures 2 and 3 provide visual comparisons of average $\textrm{RMSE}$ without latitude weighting for ensemble mean forecasts from the ECMWF ensemble and FuXi-ENS. These figures compare predictions across 3 upper-air variables ($\textrm{Z500}$, $\textrm{T850}$, and $\textrm{WS850}$) and 3 surface variables ($\textrm{MSL}$, $\textrm{T2M}$, and $\textrm{WS10M}$) at lead times of 5 days, 10 days and 15 days, using all testing data from 2018. Each figure's first and second columns depict the average $\textrm{RMSE}$ for each model, while the third column illustrates the differences between them. These $\textrm{RMSE}$ differences are represented by areas in blue, red, and white to indicate regions where FuXi-ENS performs better (lower $\textrm{RMSE}$), worse (higher $\textrm{RMSE}$), or equivalently to the ECMWF ensemble, respectively. The $\textrm{RMSE}$ spatial patterns generally show lower values in tropical versus extra-tropical regions. At a 5-day forecast lead time, FuXi-ENS typically demonstrates comparative or superior performance, as evidenced by the prevalence of blue in the difference maps. Notably, FuXi-ENS particularly outperforms ECMWF ensemble in $\textrm{WS850}$, $\textrm{WS10M}$, and $\textrm{MSL}$ over oceanic regions, and in $\textrm{T850}$ and $\textrm{T2M}$ over land areas. As lead times increase, the $\textrm{RMSE}$ contrasts diminish, with FuXi-ENS showing varied performance relative to ECMWF ensemble across different regions. This pattern suggests fluctuations in the model's efficacy over different geographic regions with increasing forecast lead times. Overall, the spatial distribution of $\textrm{RMSE}$ differences demonstrates that FuXi-ENS generally surpasses the ECMWF ensemble in accuracy at most global grid points, as indicated by the predominance of blue colors.

\FloatBarrier

\subsection{Probabilistic metrics}

In the evaluation of ensemble forecasts, statistical metrics of probability forecasts such as the $\textrm{CRPS}$ and spread-skill ratio ($\textrm{SSR}$) are indispensable. The $\textrm{CRPS}$ effectively compares the predicted probability distribution to observed values, focusing on the forecast's accuracy across its entire distribution \cite{Hersbach2000}. The $\textrm{SSR}$) assesses forecast reliability by comparing the ensemble's forecast spread with the ensemble mean error, revealing whether the ensemble's variability matches its predictive accuracy. The ensemble spread quantifies the inherent forecast uncertainty among ensemble members, with a greater spread typically suggesting higher uncertainty, which should ideally correlate with the ensemble's forecast error. This subsection compares these probabilistic metrics between the FuXi-ENS and ECMWF ensembles. 

\begin{figure}[h]
    \centering
    \includegraphics[width=\linewidth]{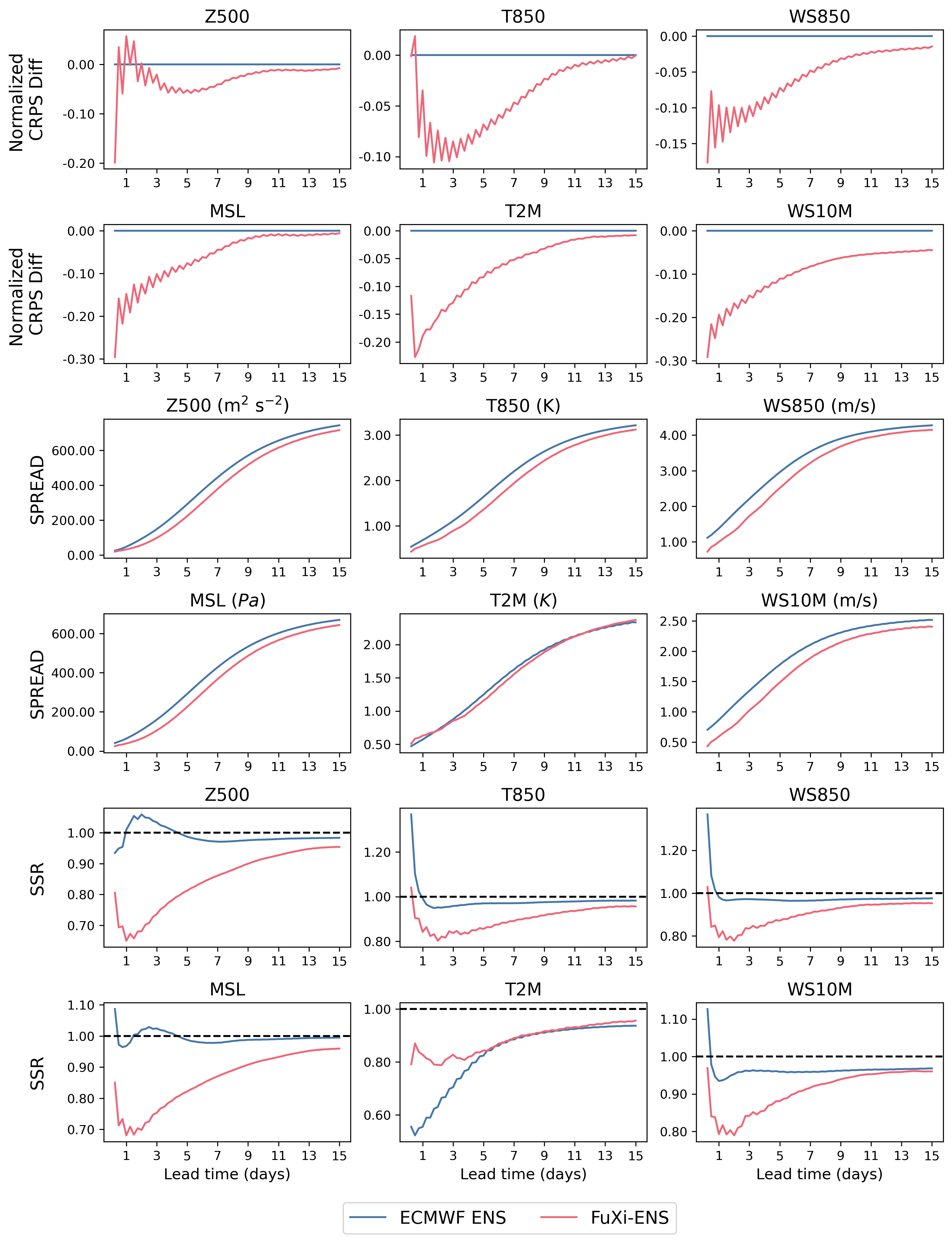}
    \caption{Comparison of normalized differences in globally-averaged, latitude-weighted $\textrm{CRPS}$ (first and second rows), ensemble spread (third and fourth rows), and $\textrm{SSR}$ (fifth and sixth rows) of ensemble forecasts from ECMWF ensemble (blue lines) and FuXi-ENS (red lines) for 6 variables: $\textrm{Z500}$, $\textrm{T850}$, $\textrm{WS850}$, $\textrm{MSL}$, $\textrm{T2M}$, and $\textrm{WS10M}$, in 15-day forecasts using testing data from 2018. The normalized differences are calculated using the ECMWF ensemble mean as the reference.}
    \label{CRPS_plot}    
\end{figure}

Figure \ref{CRPS_plot} illustrates the time series of normalized differences in globally-averaged, latitude-weighted $\textrm{CRPS}$, ensemble spread, and $\textrm{SSR}$ for the same 6 variables shown in Figure \ref{RMSE_plot}, over all 15-day forecast lead times. The CRPS values for the FuXi-ENS and ECMWF ensembles are close (refer to Supplementary Figure 4), thus only the normalized $\textrm{CRPS}$ differences are presented here.
Overall, FuXi-ENS demonstrates superior CRPS values compared to the ECMWF ensemble in 360 variable and lead time combinations ($360 = 6 \times 60$, 60 is the number of forecast steps), representing 98.1\% of these comparisons. 
Specifically, FuXi-ENS outperforms the ECMWF ensemble in $\textrm{WS850}$, $\textrm{WS10M}$, $\textrm{MSL}$, and $\textrm{T2M}$ throughout the entire 15-day forecast period. For $\textrm{Z500}$, the ECMWF ensemble initially leads in performance during the first day of forecasting but is surpassed by FuXi-ENS at about day 2. The FuXi-ENS ensemble then regains its superior performance from days 3 to 15. For $\textrm{T850}$, FuXi-ENS consistently surpasses the ECMWF ensemble over most of the forecast period except for the first 12 hours.
Additionally, the ensemble spread for FuXi-ENS and ECMWF ensemble is comparable, especially for longer forecast lead times. This contrasts with previous methods such as random perturbations (e.g., Perlin noise), which do not account for the background flow and tend to decrease after 9 days \cite{chen2023fuxi}. 
The $\textrm{SSR}$ values for ECMWF ensemble are generally close to 1 across all 6 variables over all forecast periods, indicating that the spread serves as a reliable predictor of forecast uncertainties. However, for forecasts of ECMWF within the first day, $\textrm{SSR}$ values exceed 1 for T850, WS850, MSL, and WS10M, suggesting overdispersion. In comparison, the $\textrm{SSR}$ of FuXi-ENS is closer to 1 within the first day, possibly suggesting a properly generated initial ensemble in FuXi-ENS. 
%, the $\textrm{SSR}$ values remain below 1 throughout the entire 15-day forecast period, indicating underdispersion. As lead times increase, the $\textrm{SSR}$ values for FuXi-ENS gradually converge towards 1, implying an enhancement in ensemble quality at longer forecast lead times.

Similar to Supplementary Figures 2 and 3, Supplementary Figures 5 and 6 illustrate the spatial distribution of $\textrm{CRPS}$ for ensemble forecasts by the ECMWF ensemble and FuXi-ENS across 6 variables. For $\textrm{WS850}$, $\textrm{T850}$, $\textrm{WS10M}$, and $\textrm{T2M}$, FuXi-ENS either matches or surpasses the ECMWF ensemble in $\textrm{CRPS}$ scores, as indicated by the prevalent blue and white areas. Notably, $\textrm{Z500}$ demonstrates the most significant variability in performance. These patterns suggest that FuXi-ENS generally provides more accurate probability forecasts than the ECMWF ensemble, particularly at shorter lead times and for the $\textrm{T2M}$ and $\textrm{WS10}$ variables. The spatial maps of CRPS difference maps further highlight these enhancements, reflecting FuXi-ENS's potential for improved medium-range ensemble weather predictions.

\FloatBarrier

\subsection{Extreme forecast}
Ensemble forecasts are crucial for estimating the likelihood of extreme events. In this study, we evaluate the performance of FuXi-ENS and ECMWF ensemble in predicting such events by computing the Brier score ($\textrm{BS}$) \cite{brier1950verification} for values exceeding the high and low climatological percentiles. These percentiles are determined for each variable based on the 24-year ERA5 data from 1993 to 2016 for each grid point, month of the year, and time of day.

\begin{figure}[h]
    \centering
    \includegraphics[width=\linewidth]{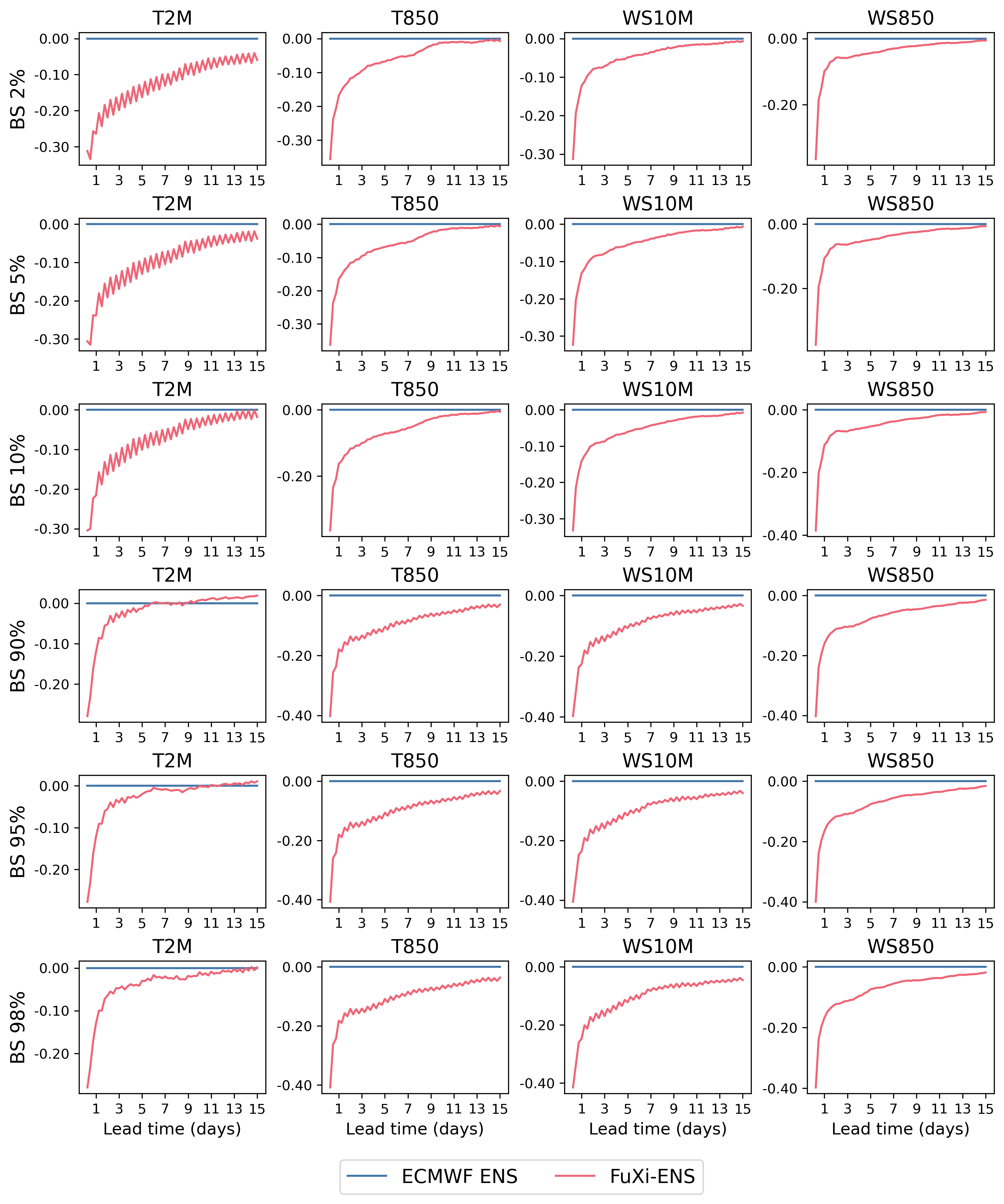}
    \caption{Comparison of the ECMWF ensemble (blue lines) and the FuXi-ENS (red lines) on the normalized differences in globally-averaged, latitude-weighted $\textrm{BS}$ for \gt90th (first row), \gt95th (second row), \gt98th (third row), \lt10th (fourth row), \lt5th (fifth row), and \lt2nd (sixth row) percentile events, (second row), and (third row) of  for 4 variables: $\textrm{Z500}$ (first column), $\textrm{T850}$ (second column), $\textrm{MSL}$ (third column), and $\textrm{T2M}$ (fourth column), in 15-day forecasts using testing data from 2018.}
    \label{BS_score}    
\end{figure}

Figure \ref{BS_score} shows a comparison of the ECMWF and FuXi-ENS ensemble's performance in terms of the normalized difference in the globally-averaged, latitude-weighted $\textrm{BS}$ for percentiles above the 90th, 95th, 98th, and below the 10th, 5th, and 2nd, for the variables $\textrm{Z500}$, $\textrm{T850}$, $\textrm{MSL}$, and $\textrm{T2M}$ across all lead times in 15-day forecasts. These thresholds represent extreme high and low events, respectively. A comprehensive analysis of 1440 cases, calculated as the product of 6 percentiles, 4 variables, and 60 forecast lead times ($1440 = 6\times 4 \times 60$), reveals that FuXi-ENS demonstrates superior performance to the ECMWF ensemble in 96.3\% of these cases. Detailed evaluations of the probabilistic TC track forecast and predictions of the record-breaking Northeast Asia heatwave are provided in the following sections.

\FloatBarrier

\subsection{Probabilistic Tropical Cyclone track forecast}

Accurate prediction of TC tracks is crucial for public safety and economic stability due to the associated high winds, heavy rainfall, and storm surges along coastlines \cite{tsai2013detection}.
Significant advancements in TC track forecasting have been made, reducing errors by approximately two-thirds over the past generation, representing a significant meteorological achievement. However, evidence suggests that TC track prediction be approaching its predictability limits. This is indicated by the analyses of linear trends in annual TC track forecast errors averaged over 5 years for the North Atlantic and eastern North Pacific \cite{Landsea2018}.
Recently, ML-based deterministic weather forecasting models \cite{bi2022panguweather,lam2022graphcast,fuxi_extreme2023} have demonstrated more accurate TC track forecasts compared to the ECMWF HRES, suggesting potentials for further improvements. Despite these advancements, TC track forecasts still involve significant uncertainties, with both conventional NWP model and ML models often failing to correctly predict landfall locations. Consequently, end users of TC forecasts require reliable track forecasts and estimates of forecast uncertainty. The traditional use of a deterministic model with static climatological errors  \cite{demaria2013improvements} proves insufficient for measuring the uncertainty of TC track forecasts. In contrast, ensemble forecasts offer substantial values by providing scenario-dependent uncertainty estimates \cite{zhang2015verification}.

\begin{figure}[h]
    \centering
    \includegraphics[width=\linewidth]{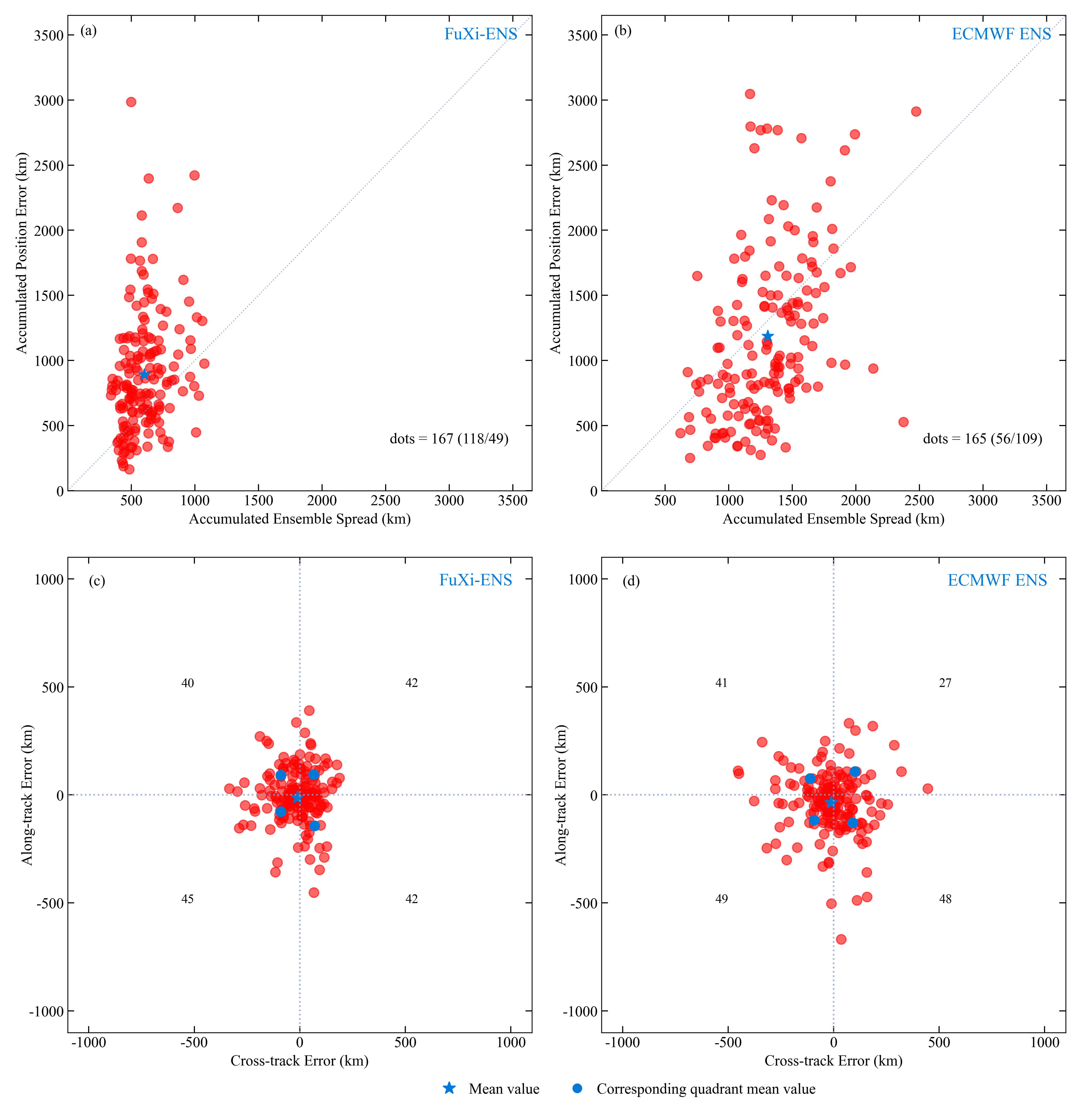}
    \caption{Scatter plots comparing $\textrm{AccERROR}_{TC}$ against $\textrm{AccSpread}_TC$, and $\textrm{AT}_{TC}$ against $\textrm{CT}_{TC}$ at a 72-hour forecast lead time for the FuXi-ENS and ECMWF ensemble. In the scatter plots of $\textrm{AccERROR}_{TC}$ against $\textrm{AccSpread}_TC$, each red dot represents an individual forecast, summarized in the format S(M/N), where S is the total number of dots, M is the number above the diagonal, and N is the number below. In the scatter plots of $\textrm{AT}_{TC}$ against $\textrm{CT}_{TC}$, red dots represent individual forecasts, with the blue dots denoting the mean values of red dots within each quadrant. The blue stars indicate the mean values of all forecasts.}
    \label{TC_ensemble_stats} 
\end{figure}

This section compares probabilistic TC track forecasts between the FuXi-ENS and the ECMWF ensemble, with the latter serving as a baseline. The ECMWF ensemble, recognized as one of the best global ensemble systems, incorporates specific enhancements for TC forecasting, such as the application of SVs to generate initial perturbations targeted at TCs \cite{puri2001ensemble}. Figure \ref{TC_ensemble_stats} presents scatter plots displaying the accumulated mean position error ($\textrm{AccERROR}_{TC}$) against the accumulated spread ($\textrm{AccSpread}_TC$) of the ensemble, as well as the along-track ($\textrm{AT}_{TC}$) component against the cross-track ($\textrm{CT}_{TC}$) component of the position errors at a 72-hour lead time. The analysis includes 20\% of all 2018 testing data, encompassing 167 predictions for FuXi-ENS and 165 for the ECMWF ensemble. Only forecasts with more than two-thirds of ensemble members detecting TCs were included in the evaluation \cite{zhang2015verification}. In the scatter plots of the $\textrm{AccERROR}_{TC}$ against the $\textrm{AccSpread}_TC$, FuXi-ENS shows 118 predictions above and 49 below the diagonal, indicating a predominance of forecasts where $\textrm{AccSpread}TC$ falls below $\textrm{AccERROR}{TC}$, thus underestimating forecast uncertainties.
In contrast, the ECMWF ensemble, with 56 predictions above and 109 below the diagonal, appears to overestimate forecast uncertainties. Overall, FuXi-ENS demonstrates superior performance, showing smaller values of both $\textrm{AccSpread}_TC$ and $\textrm{AccERROR}_{TC}$ compared to the ECMWF ensemble, indicating more accurate track forecasting.

The analysis of scatter plots comparing $\textrm{AT}_{TC}$ and $\textrm{CT}_{TC}$ provides insights into the TC speed (slow/fast) and directional biases (left/right) relative to the best track. In these plots, mean values are depicted by blue stars, while the blue dots represent the mean values within each quadrant. For the ECMWF ensemble, the forecast distribution across the four quadrants is as follows: 27 in the upper right, 48 in the lower right, 41 in the upper left, and 49 in the lower left.
This distribution suggests that the mean track generally lags behind the best track (41 + 27 $\gt$ 49 + 48), as evidenced by the higher numbers in the lower quadrants (97) versus the upper quadrants (68). Furthermore, the ECMWF ensemble's predictions predominantly occur to the left of the best track, as indicated by the predominance of cases in the left quadrants compared to the right quadrants (41 + 49 $\gt$ 27 + 48).
In contrast, the FuXi-ENS shows a more balanced distribution with 42 forecasts each in the upper right and lower right quadrants, 40 and 45 in the upper left and lower left quadrants, respectively.
This pattern shows only a slight lag behind the best track (40 + 42 $\lt$ 45 + 42), and no significant deviation in the perpendicular direction (40 + 45 $\approx$ 42 + 42). In summary, the the FuXi-ENS shows relatively smaller mean position errors compared to the ECMWF ensemble, thereby demonstrating its superior performance in TC track forecasting.
Additionally, Supplementary Figure 11 presents box plots of $\textrm{AccERROR}_{TC}$, $\textrm{AccSpread}_TC$, $\textrm{AT}_{TC}$, and $\textrm{CT}_{TC}$ in 5-day forecasts for the FuXi-ENS and ECMWF ensemble. This figure clearly shows that, throughout the 5-day forecast period, the ECMWF ensemble consistently has a higher accumulated ensemble spread and positional error compared to the FuXi-ENS. Both $\textrm{AT}_{TC}$ and $\textrm{CT}_{TC}$ exhibit comparable magnitudes, though $\textrm{AT}_{TC}$ in the ECMWF ensemble shows significant deviations.
Moreover, a case study on predicting the 2018 Hurricane Florence is presented in Supplementary material.

\FloatBarrier

\subsection{Prediction of the Record-Breaking Northeast Asia heatwave in 2018}

Heat waves are typically defined as prolonged periods of extreme heat resulting from consecutive hot days, which have negative impacts on human health, ecosystems, agriculture, and infrastructure. During the summer of 2018, Northeast Asia experienced a record-breaking heatwave, affecting various regions in Japan, the Korean Peninsula, and northeastern China \cite{WMO2018,Pang2020}, resulting in considerable economic damage and a significant number of fatalities. The Japan Meteorological Agency (JMA) reported a recording-setting maximum temperature of 41.1 °C in Kumagaya on July 23, the highest ever observed in the country. Similarly, on August 1, the Korean Meteorological Administration (KMA) recorded a peak temperature of 39.6°C in Seoul, the hottest day in 111 years. In northeastern China, Liaoning and Jilin provinces experienced temperatures approaching 39°C during July and August 2018, as reported by the China Meteorological Administration (CMA). The heatwave resulted in at least 138  in Japan and 42 in South Korea, with over 7,000 people hospitalized in Japan and 3,000 in South Korea due to heat-related illnesses. Given these severe consequences, improving the accuracy of forecasts for such high-impact weather events is critical to provide timely and effective warnings and mitigate their devastating effects.

\begin{figure}[h]
    \centering
    \includegraphics[width=\linewidth]{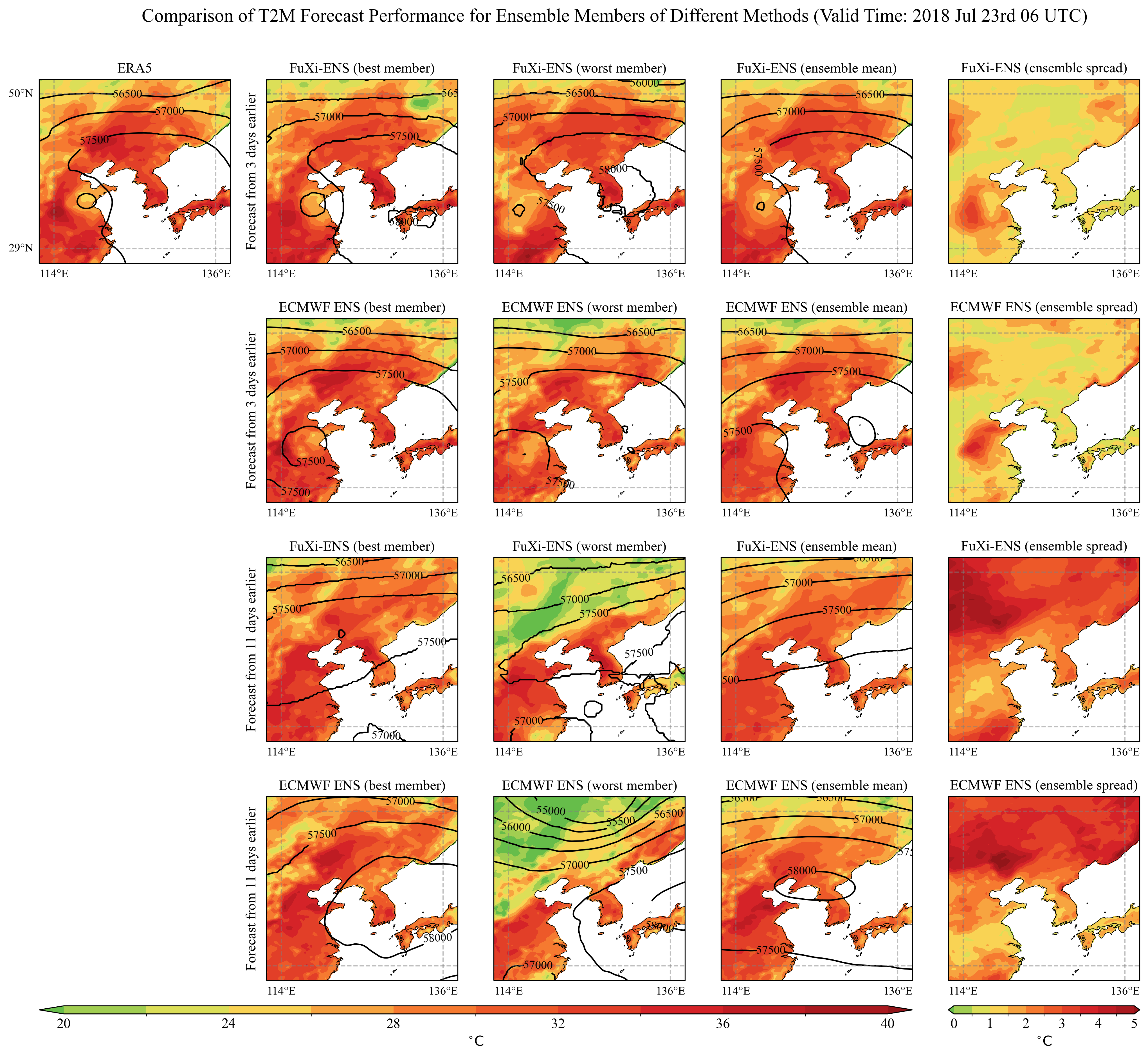}
    \caption{Spatial distributions of $\textrm{T2M}$ generated by FuXi-ENS and ECMWF ensemble during the 2018 Northeast Asia heatwave. The first column displays the ERA5 reanalysis data for 6 UTC on July 23, 2018. Columns 2 through 5 are predictions from FuXi-ENS (first and third rows) and ECMWF ensemble (second and fourth rows), showing the best member (second column), worst member (third column), ensemble mean (fourth column), and ensemble spread (fifth column). The top two rows are forecasts made from 3 days earlier, and the bottom two rows are forecasts made from 11 days earlier. The black contours indicate the $\textrm{Z500}$.}
    \label{heatwave_T2M} 
\end{figure}

Figure \ref{heatwave_T2M} presents a comparative analysis of T2M forecasts generated by the FuXi-ENS and ECMWF ensembles during the 2018 Northeast Asia heatwave. The ERA5 reanalysis data, shown in the first column, serves as a benchmark, depicting the spatial distribution of T2M at 6 UTC on July 23, 2018. Subsequent columns illustrate forecasts from both ensembles, including the best-performing member, the worst-performing member, the ensemble mean, and the ensemble spread. The rankings of different ensemble members are based on the T2M RMSE averaged over the depicted land area. Both ensemble model demonstrate comparable performances across the best and worst members, as well as the ensemble mean and spread. The forecasts include two lead times: 3 days (top two rows) and 11 days (bottom two rows) prior to the event. As the initialization date approaches the event, both models exhibit a reduced ensemble spread, indicating reduced forecast certainty. The spatial patterns of ensemble spread within the FuXi-ENS are similar to those of the ECMWF ensemble, reflecting consistency in estimating forecast uncertainties. 

Several studies attribute the extreme heatwave to the positive anomalies in geopotential height (persistent high pressure) over Northeast Asia, which induced anomalous
downward motion, amplified solar radiation, and consequently raised local temperatures \cite{xu2019large,lu2020unusual,Pang2020,ren2020attribution}. The Z500 contours, superimposed on the forecasts in Figure \ref{heatwave_T2M}, demonstrate that both the FuXi-ENS and ECMWF ensembles accurately predict the large-scale circulation patterns for forecasts initialized 3 days before the event, closely aligning with the ERA5 dataset.

\FloatBarrier

%\subsection{Comparison of different ensemble generation methods for ML models}
%train a model to generate initial condition perturbation to maximize ensemble spread similar to the singular-vector approach at ECMWF, which maximizes the growth of ensemble 

\section{Discussion} 

In recent years, ML models have demonstrated significant potential in weather forecasting, surpassing the performance of world's top NWP models in medium-range forecasting. However, ML models face a greater challenge in outperforming NWP models in ensemble forecasting, which is critical for estimating the likelihood of extreme weather events. The accuracy of weather forecasts decreases with longer lead times due to the inherently chaotic nature of weather systems, which amplifies uncertainty as lead times increase. Unlike deterministic forecasts, which fail to capture these uncertainties, ensemble forecasts allow for assessment of the uncertainty in predictions by analyzing the spread of ensemble members. Moreover, the increasing intensity and frequency of extreme weather events, such as tropical cyclones (TCs), due to climate change, makes accurate uncertainty estimates increasingly important.

Recent ML models, such as Google's GenCast and SEEDS model, rely on either EDA or two operational NWP ensemble members for their forecasts. This reliance complicates their implementation in operational settings and requires additional input data. Despite their superior forecasting performance compared to ECMWF ensemble forecasts, the 2\textdegree spatial resolution of SEEDS limits its utility for a wide range of applications. Furthermore, as a diffusion model, GenCast requires 20 solver steps to produce a single-step forecast, making it relatively time-consuming compared to other ML models.
This paper introduces FuXi-ENS, a ML-based medium-range ensemble weather forecasting model that demonstrates superior performance compared to the ECMWF ensemble at a spatial resolution of 0.25\textdegree. FuXi-ENS is capable of generating 6-hourly forecasts up to 15 days, including 5 upper-air atmospheric variables at 13 pressure levels and 13 surface variables. A unique feature of the FuXi-ENS model is its incorporation of a innovative combination of CRPS and KL loss, which outperforms the conventional use of L1 loss combined with KL loss in classical VAE models. The flow-dependent perturbations of the FuXi-ENS significantly enhance its ensemble forecast performance. These perturbations are introduced at both the initial conditions and each forecast step, similar to traditional ensemble models.
If we should have seen further, it is by standing upon the shoulders of giants.
Results demonstrate that FuXi-ENS outperforms ECMWF ensemble in 98.1$\%$ of 360 variable and forecast lead time combinations in terms of CRPS.
Furthermore, case studies of high-impact weather events to further evaluate FuXi-ENS's performance, such as probabilistic TC track forecasts and 2018 Northeast Asia heatwave.
Compared to conventional NWP-based ensemble forecasts, the FuXi-ENS model distinguishes itself  for its ability to rapidly and efficiently generate extensive ensemble forecasts, requiring significantly less time and computational resources. It can generate a single member of 15-day forecasts with a temporal resolution of 6 hours in approximately 10 seconds using an Nvidia A100 GPU.

Beyond medium-range weather forecasting, the framework used in developing FuXi-ENS shows promise for applications in other critical forecast scenarios that require ensemble predictions, such as subseasonal-to-seasonal forecasts, seasonal forecasts, and wave forecasts. The role of ML models in advancing weather forecasting is expected to grow, particularly for extreme weather events where uncertainty quantification is essential. Furthermore, ensemble forecasting plays a crucial role in data assimilation, as ensemble-based data assimilation methods can generate more accurate and ﬂow-dependent estimates of background-error covariances \cite{bonavita2016evolution}, thereby improving the quality of initial conditions. However, computational constraints often limit ensemble sizes significantly below the model's degrees of freedom \cite{houtekamer2016review}, leading to sampling errors in the ensemble-based data assimilation approach. FuXi-ENS mitigates these issues by reducing the cost of ensemble forecasting, facilitating the generation of larger ensembles for improved data assimilation.

\section{Methods} 
 
\subsection{Data}
ERA5 stands as the fifth iteration of the ECMWF reanalysis dataset, offering a rich array of surface and upper-air variables. It operates at a remarkable temporal resolution of 1 hour and a horizontal resolution of approximately 31 km, covering data from January 1950 to the present day \cite{hersbach2020era5}. Recognized for its expansive temporal and spatial coverage coupled with exceptional accuracy, ERA5 stands as the most comprehensive and precise reanalysis archive globally. In our study, we utilize 6-hourly ERA5 dataset a spatial resolution of $0.25^\circ$ ($721\times1440$ latitude-longitude grid points). 

The FuXi-ENS model forecasts a total of 78 variables, encompassing 5 upper-air atmospheric variables across 13 pressure levels (50, 100, 150, 200, 250, 300, 400, 500, 600, 700, 850, 925, and 1000 hPa), and 11 surface variables. Among the upper-air atmospheric variables are geopotential (${\textrm{Z}}$), temperature (${\textrm{T}}$), u component of wind (${\textrm{U}}$), v component of wind (${\textrm{V}}$), and specific humidity (${\textrm{Q}}$). The surface variables include 2-meter temperature (${\textrm{T2M}}$), 2-meter dewpoint temperature (${\textrm{D2M}}$), sea surface temperature (${\textrm{SST}}$), 10-meter u wind component (${\textrm{U10}}$), 10-meter v wind component (${\textrm{V10}}$), 100-meter u wind component (${\textrm{U100}}$), 100-meter v wind component (${\textrm{V100}}$), mean sea-level pressure (${\textrm{MSL}}$), surface net solar radiation (${\textrm{SNSR}}$), surface net solar radiation downwards (${\textrm{SSRD}}$), total sky direct solar radiation at surface (${\textrm{FDIR}}$), top net thermal radiation (${\textrm{TTR}}$) \footnote{${\textrm{TTR}}$, known as the negative of outgoing longwave radiation (${\textrm{OLR}}$)}, and ${\textrm{TP}}$. Table \ref{glossary} presents a detailed list of variables and their corresponding abbreviations. Variables such as ${\textrm{U100}}$, ${\textrm{V100}}$, ${\textrm{SSR}}$, ${\textrm{SSRD}}$, and ${\textrm{FDIR}}$ have been selected for their relevance and utility in the wind and solar energy forecasting industries.

The model's training relies on 15 years of data spanning from 2002 to 2016, 1 year (2017) data for validation and 1 year (2018) data for testing. More detailed evaluations of ${\textrm{TP}}$ and  predictions for the year 2022 can be found in the supplementary material.
%The validation set contains data from year 2002 to 2016, which is used for calculating climatology of forecast data.

Since the ECMWF EPS became fully operational in 1992, it has established as a global leader in medium-range and subseasonal-to-seasonal ensemble forecasting. The ECMWF ensemble consists of 51 forecasts, which include one control forecast generated from the optimal estimate of initial conditions, alongside 50 perturbed forecasts, each generated with slight variations in both initial conditions and model physics. On June 27, 2023, the spatial resolution of the ECMWF ensemble was upgraded to 9 km. For our analysis, we leveraged archived ECMWF ensemble forecasts from 2018, which has a spatial resolution of $0.25^\circ$.
Furthermore, we validated that the continuous ranked probability score (CRPS) based on all 51 members. To compute the $\textrm{BS}$, all 51 members of the ECMWF ensemble forecasts are required. Due to the substantial data volume, we retrieved the full ensemble every five days throughout 2018. Consequently, 20\% samples in 2018 are available for $\textrm{BS}$ calculation compared to other metrics.
%Given the substantial volume of data associated with all 51 members, we chose to download mean and standard deviation values from ECMWF archive for the entire 2018. Furthermore, we validated that the continuous ranked probability score (CRPS), calculated either based on all 51 members or their mean and standard deviation, yields nearly identical results. To compute the $\textrm{BS}$, all 51 members of the ECMWF ensemble forecasts are required. Due to the substantial data volume, we retrieved the full ensemble every five days throughout 2018. Consequently, 20\% samples in 2018 are available for $\textrm{BS}$ calculation compared to other metrics.

\begin{table}
\centering
\caption{\label{glossary} A summary of all the input and output variables. The "Type" indicates whether the variable is a time-varying variable including upper-air, surface, and geographical variables, or a temporal variable. The “Full name” and “Abbreviation” columns refer to the complete name of each variables and their corresponding abbreviations in this paper. The "Role" column clarifies whether each variable serves as both an input and a output, or is solely utilized as an input by our model.}
\begin{tabularx}{\textwidth}{cXcc}
\hline
\textbf{Type} & \textbf{Full name} & \textbf{Abbreviation} & \textbf{Role} \\
\hline
upper-air variables & geopotential & ${\textrm{Z}}$ & Input and Output   \\
                    & temperature  & ${\textrm{T}}$ & Input and Output   \\
                    & u component of wind & ${\textrm{U}}$ & Input and Output  \\
                    & v component of wind & ${\textrm{V}}$ & Input and Output  \\
                    & specific humidity & ${\textrm{Q}}$ & Input and Output  \\
                    \hline
surface variables   & 2-meter temperature & ${\textrm{T2M}}$ & Input and Output  \\
                    & 2-meter dewpoint temperature & ${\textrm{D2M}}$ & Input and Output  \\
                    & sea surface temperature & ${\textrm{SST}}$ & Input and Output  \\
                    & 10-meter u wind component & ${\textrm{U10}}$ & Input and Output  \\
                    & 10-meter v wind component & ${\textrm{V10}}$ & Input and Output  \\
                    & 100-meter u wind component & ${\textrm{U100}}$ & Input and Output  \\
                    & 100-meter v wind component & ${\textrm{V100}}$ & Input and Output  \\ 
                    & mean sea-level pressure & ${\textrm{MSL}}$ & Input and Output  \\
                    & surface net solar radiation & ${\textrm{SNSR}}$ & Input and Output \\
                    & surface net solar radiation downwards & ${\textrm{SSRD}}$ & Input and Output \\
                    & total sky direct solar radiation at surface & ${\textrm{FDIR}}$ & Input and Output \\
                    & top net thermal radiation & ${\textrm{TTR}}$ & Input and Output \\
                    & total precipitation & ${\textrm{TP}}$ & Input and Output \\
\hline
geographical        & orography & ${\textrm{OR}}$ & Input \\
                    & land-sea mask & ${\textrm{LSM}}$ & Input \\
                    & latitude & ${\textrm{LAT}}$ & Input \\
                    & longitude & ${\textrm{LON}}$ & Input \\  
\hline                  
temporal            & hour of day & ${\textrm{HOUR}}$ & Input \\
                    & day of year & ${\textrm{DOY}}$ & Input \\
                    & step & ${\textrm{STEP}}$ & Input \\                    
\hline
\end{tabularx}
\end{table}

We assessed tropical cyclone (TC) forecasts using the International Best Track Archive for Climate Stewardship (IBTrACS) \cite{knapp2010,knapp2018} dataset from the National Oceanic and Atmospheric Administration (NOAA), as the reference. The IBTrACS aggregates global best track data into a comprehensive archive, in which each track represents a TC's position at six-hourly intervals using latitude and longitude coordinates, among other critical features. Additionally, we employed a modified version of the ECMWF TC tracker algorithm (refer to Subsection \ref{TC_tracking}), as detailed by van der Grijn \cite{der2002tropical} to both the ECMWF and FuXi-ENS ensemble members, as well as to the ERA5 dataset, facilitating the extraction and evaluation of TC track forecasts. Statistically robust characteristics only emerge from sufficiently large sample sizes (ensemble numbers), thus we excluded forecasts with fewer than two-thirds of ensemble members detecting TCs for evaluating probabilistic TC track forecasts \cite{zhang2015verification}.

\subsection{FuXi-ENS model}

Previous ML-based medium-range weather forecasting models have predominantly employed deterministic encoder-decoder architectures \cite{bi2022panguweather,lam2022graphcast,chen2023fuxi,olivetti2023advances}. These models, aimed at minimizing the root mean square error ($\textrm{RMSE}$) between their outputs and ERA5 reanalysis data, fail to provide the uncertainty inherent in their forecasts. This limitation is particularly critical when predicting high-impact extreme weather events. Moreover, unlike deterministic models that use ERA5 data as a benchmark, ensemble forecasts lack a definitive "ground truth". Consequently, direct comparisons between ensemble model outputs and actual conditions are impractical, posing a significant challenge for the development of supervised ML models in ensemble weather forecasting.

Building on previous research into ensemble models using conventional NWP models, our approach introduces perturbations at both initial conditions and each forecast step to account for both initial condition errors and model uncertainties. We develop an auto-regressive ML model, FuXi-ENS, for ensemble weather forecasting that can address the limitations of traditional deterministic encoder-decoder models. This model incorporates a Variational Autoencoder (VAE) \cite{doersch2016tutorial,zhao2017learning,kingma2019introduction} due to its probabilistic nature and capability to quantify uncertainty, crucial for ensemble forecasting. VAEs are based on a probabilistic generative model. In a VAE, the encoder transforms input data into a probability distribution, typically Gaussian, from which the perturbations are sampled. This sampling process introduces randomness and variability, providing a direct measure of uncertainty. The distribution's spread reflects the model's uncertainty about the latent representation of input data.
%XH, 因为集合预报没有加多少扰动这种gt可以学习，所以我们通过优化集合预报中的crps指标来优化模型。
%这里在解释下为什么vae这种生成式模型可以提取不确定性，因为robust什么的

\begin{figure}[h]
    \centering
    \includegraphics[width=\linewidth]{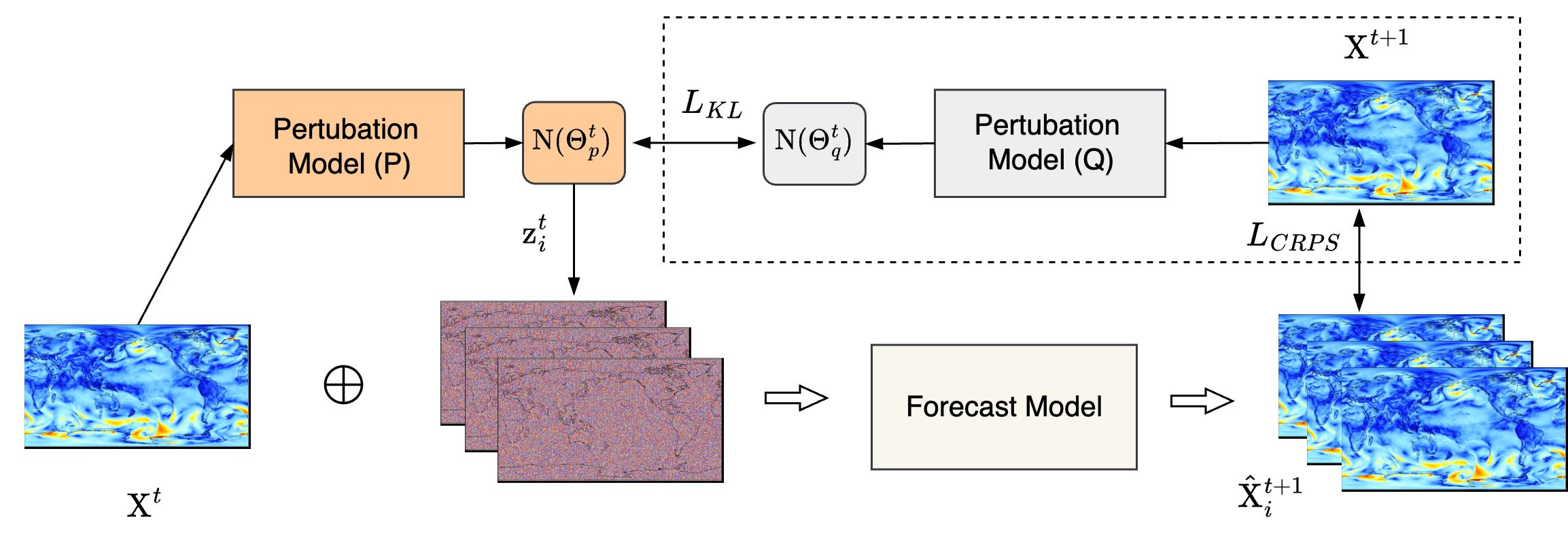}
    \caption{Schematic diagram of the structures of the FuXi-ENS model.}
    \label{model}    
\end{figure}

The FuXi-ENS model consists of two main components: a perturbation model and a forecasting model (see Figure \ref{model}). The perturbation model, based on a VAE, transforms input data into a Gaussian distribution with the same dimensions as the input, thereby reflecting the probabilistic characteristics of the input data. In contrast, the forecasting model, which uses an encoder-decoder framework, generates predictions from the perturbed initial conditions sampled from this Gaussian distribution. This method captures the inherent uncertainty in the data and facilitates the generation of ensemble forecasts through repeated sampling. The number of samples equals the number of ensemble members produced. This process is repeated after each forecasting step, enhancing forecast uncertainty estimation and the model's utility in ensemble forecasting. To enhance comprehension, we can draw an analogy between this ML methodology and established techniques in traditional ensemble forecasting. Within our framework, the deterministic forecasts rely on the forecasting model and initial conditions, while the perturbation model introduces flow-dependent variations to account for uncertainties in the initial conditions and model predictions at each forecasting step.

The two main components of the FuXi-ENS model are the perturbation model and the forecasting model. The model takes a input data cube with dimensions of $2\times78\times721\times1440$, representing meteorological variables from two previous time steps (${t-1}$ and ${t}$), the count of upper-air and surface variables (${C}$), and the number of grid points along latitude (${\textrm{H}}$) and longitude (${\textrm{W}}$), respectively. This data cube is initially reshaped to dimensions of $(2C)\times H \times W$ ($156\times721\times1440$). The perturbation model starts by reducing the dimensions of the meteorological data cube to $2C\times90\times180$ through a 2-dimensional (2D) convolution layer with a kernel size of 8 and a stride of 8. Simultaneously, geographical and temporal variables are processed by an identical 2D convolution layer. The outputs are then concatenated with the meteorological data, and further processed through 14 consecutive Swin Transformer \cite{liu2021swin} blocks, each employing a $18 \times 18$ window. The data is subsequently restored to its original dimensions via a 2D transposed convolution layer, also with a kernel size and stride of 8 \cite{Zeiler2010}. The perturbation model finally generates a Gaussian distribution (${\textrm{N($\Theta$}^t}_p)$), defined by a mean matrix $\mu^t$ and a covariance matrix $\sigma^t$ with dimensions $156\times721\times1440$. The forecasting model processes the perturbed initial conditions ($\tilde{\textrm{\textbf{X}}}^t = \textrm{\textbf{X}}^t + \textrm{\textbf{z}}^t$) beginning with a $8 \times 8$ 2D convolution layer, followed by 40 transformer blocks, and a $8 \times 8$ 2D transposed convolution layer, producing the final ensemble forecasts $\hat{\textrm{\textbf{X}}}^{t+1}$. The ensemble size equals the number of samples drawn from the Gaussian distribution ${\textrm{N($\Theta$}^t}_p)$.

Training FuXi-ENS focuses minimizing the CRPS loss, a key metric for assessing the quality of ensemble forecasts. A regularization term is used to align the model-derived Gaussian distribution (${\textrm{N($\Theta$}^t}_p)$) with the target distribution, essential for accurately representing prediction uncertainty. A significant challenge is reconciling discrepancies between these distributions, mainly due to prediction errors. Knowledge distillation is employed in transferring information from target to predicted distributions. Key to this approach is the perturbation model ($\textrm{Q}$), which converts target data into a Gaussian distribution that supervises the perturbation model $\textrm{P}$ to minimize the Kullback–Leibler (KL) divergence loss ($\textrm{L}_{\textrm{KL}}$). This loss quantifies discrepancies between the two models' distributions. During training, perturbation model $\textrm{Q}$ processes target data from two previous time steps: $\textrm{\textbf{X}}^t$ and $\textrm{\textbf{X}}^{t+1}$. It predicts a Gaussian distribution (${\textrm{N($\Theta$}^t}_q)$) similar to that of model $\textrm{P}$. Sampling of intermediate perturbation vectors occurs from model $\textrm{Q}$'s distribution during training, and from model $\textrm{P}$'s distribution (${\textrm{N($\Theta$}^t}_p)$) during testing. Moreover, the CRPS loss is calculated between the ensemble forecast ( $\hat{\textrm{\textbf{X}}}^{t+1}$) and the target data ${\textrm{\textbf{X}}}^{t+1}$. The overall loss function is defined as:
\begin{equation}
\label{L_CRPS}
    \textrm{L} = \textrm{L}_{\textrm{CRPS}}(\hat{\textrm{\textbf{X}}}^{t+1}_i, \textrm{\textbf{X}}^{t+1}_i) + \lambda\textrm{L}_{\textrm{KL}}({\textrm{N($\Theta$}^t}_p, {\textrm{N($\Theta$}^t}_q))
\end{equation}
where $\lambda$ is a tunable coefficient, set to $1\times10^{-4}$, balancing $\textrm{L}_{KL}$ and $\textrm{L}_{\textrm{CRPS}}$ loss terms. This design ensures that perturbation vectors closely approximate the true data distribution and improve the CRPS of the ensemble forecasts. During training, each GPU generates one ensemble member, resulting in a total of eight members.

In this study, we utilize 48 ensemble members for medium-range ensemble forecasting. As illustrated in Supplementary Figure 14, the FuXi-ENS model incorporating CRPS in the loss function outperforms the same model utilizing L1 loss. The loss function for the FuXi-ENS model, which is trained with a combination of L1 and KL loss, is defined as follows:
\begin{equation}
\label{L1_loss}
    \textrm{L} = \textrm{L1}(\hat{\textrm{\textbf{X}}}^{t+1}, \textrm{\textbf{X}}^{t+1}) + \lambda\textrm{L}_{\textrm{KL}}(\textrm{P}^t, \textrm{Q}^t)
\end{equation}
where $\textrm{L1}$ replaces the $\textrm{CRPS}$ loss.

\FloatBarrier

The FuXi-ENS model is developed using the Pytorch framework \citep{Paszke2017}. It adopts an autoregressive training regime and a curriculum training schedule \cite{lam2022graphcast}, wherein the number of the autoregressive steps from 1 to 3. Each of these steps consists of 3,000 training iterations. A batch size of 1 was used per GPU during training, which was conducted on a cluster of 8 Nvidia A100 GPUs. For optimization purposes, the AdamW optimizer \cite{kingma2017adam,loshchilov2017decoupled} was employed, configured with parameters ${\beta_{1}}$=0.9 and ${\beta_{2}}$=0.95, an initial learning rate of 2.5$\times$10$^{-4}$, and a weight decay coefficient of 0.1.

\subsection{Tropical cyclone tracking method}
\label{TC_tracking}

In our study, we implemented a modified version of the ECMWF TC tracking algorithm, as detailed by Zhong et al. \cite{fuxi_extreme2023}, applied to both ECMWF ensemble and FuXi-ENS forecasts. As noted by Zhong et al. \cite{fuxi_extreme2023}, multiple crucial adjustments to the algorithm aimed at reducing false TC tracks in the ML model predictions were made.

The ECMWF TC tracker begins by using TC observation data to estimate the initial position of the TC. The tracker's primary function is to identify the cyclone within the initial conditions (analysis) by locating the minimum $\textrm{MSL}$ within a 445 km radius of the observed TC location. We enhanced detection by incorporating local maxima of vorticity at 10 meters to increase the number of TC center candidates. The tracker subsequently monitor these candidates in forecasts until the cyclone becomes undetectable. It updates the TC position by calculating a displacement based on a linear extrapolation of locations from previous two steps, and an advection vector averaged from zonal ($\textrm{U}$) and meridional ($\textrm{V}$) wind components across multiple pressure levels, scaled by the forecast time step of 6 hours. 

In the first forecast step, where the lead time is 6 hours and only one location estimate is available, linear extrapolation is not possible, necessitating reliance solely on the advection vector. Subsequent estimates involves assessing all local $\textrm{MSL}$ minima within a 445 km radius. The inherent smoothness of machine learning (ML)-based forecasts and the minor differences among adjacent points within ML forecasts can lead to in false local minima, promoting us to decrease the spatial resolution of FuXi-ENS and ECMWF ensemble forecasts from 0.25\textdegree to 1.25\textdegree by average-pooling. TC Center candidates are then selected from the average-pooled values positioned at grid's center. The search for the nearest candidate minima that meet three specified conditions (see Supplementary Table 1) continues unless the elevation exceeds 1000 meters, indicating no cyclone presence. Upon the selection of center candidates, the data is restored to its original resolution, with final candidate needing to fulfill an additional criterion involving the local minima of $\textrm{Z850}$. 
To enhance tracking accuracy and prevent erroneous large displacements, we considered limiting the displacement to no more than 3 times the distance moved from the previous time step. However, we chose not to restrict the angle changes between directions.

\subsection{Evaluation method}
%RMSE, （ACC）
%CRPS, SSR, rank histogram
%BS 一般认为集合成员数增加有利于极端天气预报
In this work, we assess the performance of both FuXi-ENS and ECMWF ensemble using their respective initial conditions for the year 2018, focusing on 00 and 12 UTC initialization times. Specifically, we utilize the ensemble control forecast (ENS-fc0) to evaluate ECMWF ensemble, and the ERA5 dataset to evaluate FuXi-ENS. Following the methodology outlined by Rasp et al. \cite{rasp2020weatherbench,rasp2023weatherbench}, our evaluation metrics include $\textrm{RMSE}$, $\textrm{ACC}$, CRPS \cite{Hersbach2000,Sloughter2010}, ensemble spread, and spread-skill ratio ($\textrm{SSR}$), and Brier Score ($\textrm{BS}$) \cite{brier1950verification}. Specifically, we analyze the deterministic forecast performance of the ensemble mean through the latitude-weighted $\textrm{RMSE}$ and $\textrm{ACC}$, calculated as follows:

\begin{equation}
    \textrm{RMSE}(c, \tau) =\frac{1}{\mid D \mid}\sum_{t_0 \in D} \sqrt{\frac{1}{H \times W} \sum_{i=1}^H\sum_{j=1}^{W} a_i {( \bar{\hat{X}}^{t_0 +\tau}_{c,i,j} - X^{t_0 +\tau}_{c,i,j} )}^{2}}
\end{equation}

\begin{equation}
    \textrm{ACC}(c, \tau) = \frac{1}{\mid D \mid}\sum_{t_0 \in D} \frac{\sum_{i, j} a_i (\bar{\hat{X}}^{t_0 +\tau}_{c,i,j} - M^{t_0 +\tau}_{c,i,j}) (\bar{X}^{t_0 +\tau}_{c,i,j} - M^{t_0 +\tau}_{c,i,j})} {\sqrt{ \sum_{i, j} a_i (\bar{\hat{X}}^{t_0 +\tau}_{c,i,j} - M^{t_0 +\tau}_{c,i,j})^2 \sum_{i, j} a_i(\bar{X}^{t_0 +\tau}_{c,i,j} - M^{t_0 +\tau}_{c,i,j})^2}}
\end{equation}
where ${t_0}$ denotes the forecast initialization time in the testing set ${D}$, while ${\tau}$ represents the forecast lead time steps after ${t_0}$. The climatological mean, denoted as ${\textbf{M}}$, is derived from ERA5 reanalysis data spanning from 1993 to 2016. To enhance the differentiation of forecast performance among models with minor differences, we use the normalized $\textrm{RMSE}$ difference between model A and baseline B, computed as \((\textrm{RMSE}_A-\textrm{RMSE}_B)/\textrm{RMSE}_B\). Similarly, the normalized $\textrm{ACC}$ difference is computed as \((\textrm{ACC}_A-\textrm{ACC}_B)/(1-\textrm{ACC}_B)\). Negative values in normalized $\textrm{RMSE}$ difference and positive values in normalized \(\textrm{ACC}\) difference suggest superior performance of model A compared to model B.

In addition, we evaluate ensemble forecasts using various metrics, including the globally-averaged, latitude-weighted $\textrm{CRPS}$, ensemble spread, $\textrm{SSR}$. The $\textrm{CRPS}$ is determined using the following equation:
\begin{equation}
\textrm{CRPS} = a_i \int_{-\infty}^{\infty} [F(\hat{\textbf{X}}^{t_0 +\tau}_{c,i,j})-\mathcal{H}(\textbf{X}^{t_0 +\tau}_{c,i,j} \le z)] \,dz \
\end{equation}
where $F$ denotes the cumulative distribution function (CDF) of the variable ($\hat{\textbf{X}}^{t_0 +\tau}_{c,i,j}$), and $\mathcal{H}$ is an indicator function that equals 1 when $\textbf{X}^{t_0 +\tau}_{c,i,j} \le z$ is true, and 0 otherwise \cite{Wilks2011}. In deterministic forecasts, the $\textrm{CRPS}$ is equal to the mean absolute error ($\textrm{MAE}$) \cite{Hersbach2000}. The calculation of normalized difference in $\textrm{CRPS}$ follows the same procedure to that of the normalized difference in $\textrm{RMSE}$.

On the other hand, ensemble forecasts are designed to capture the full range of uncertainty. The $\textrm{SSR}$ is a pivotal metric defined as the ratio of the ensemble spread (standard deviation) to the forecast skill ($\textrm{RMSE}$). This ratio measures the alignment between the ensemble spread and $\textrm{RMSE}$ of the ensemble mean, thus serving as a essential indicator of the forecast's reliability. Palmer et al. \cite{palmer2006ensemble} demonstrated that in a perfect ensemble, the mean spread match the $\textrm{RMSE}$ over the same period, suggesting that the ensemble spread can reliably predict the error in the ensemble mean. The formula for calculating the globally-averaged, latitude-weighted ensemble spread is:
\begin{equation}
\label{spread_equation}
    \textrm{Spread}(c, \tau) = \frac{1}{\mid D \mid}\sum_{t_0 \in D} \sqrt{\frac{1}{H \times W} \sum_{i=1}^H\sum_{j=1}^{W} a_i var( \hat{\textbf{X}}^{t_0 +\tau}_{c,i,j} )}
\end{equation}
where $var( \hat{\textbf{X}}^{t_0 +\tau}_{c,i,j} )$ denotes the variance within the ensemble for each grid point $(i,j)$ at lead time $\tau$ from the initial time $t_0$. A $\textrm{SSR}$ value of 1 suggests a reliable ensemble according to Fortin et al. \cite{Fortin2014}. $\textrm{SSR}$ values below one indicate that the ensemble is underdispersive, suggesting insufficient variability, whereas values exceeding one indicate overdispersion, pointing to excessive variability.

To evaluate the performance of extreme weather forecasts, we use the globally-averaged, latitude-weighted $\textrm{BS}$, computed as follows:
\begin{equation} 
\label{BS_equation}
     \textrm{BS} = a_i \overline{(p_{forecast} - o)^2}
\end{equation}
where $p_{forecast}$ represents the model's forecast probability of an event, defined by whether the ensemble members exceed or fall below a specified threshold $T$ within an $N$-member ensemble. Specifically, $p_{forecast} = \frac{1}{N}\sum_{n}[X^n > T]$ for events above the threshold, and $p_{forecast} = \frac{1}{N}\sum_{n}[X^n < T]$ for events below it. The $\textrm{BS}$ quantifies the mean squared difference between these predicted probabilities ($p_{forecast}$) and the observed outcomes ($o$), which are either 0 or 1. Similar to $\textrm{RMSE}$, the $\textrm{BS}$ is negatively orientated, with lower values indicating more accurate forecasts, ranging from 0 for perfect forecasts to 1 for consistently incorrect forecasts. In this study, the $\textrm{BS}$ is calculated using a range of climatological percentiles as thresholds to define extreme events: \gt90th, \gt95th, and \gt98th for extreme high events, while \lt10th, \lt5th, and \lt2nd for extreme low events.

To evaluate the accuracy of ensemble TC track forecasts, ensemble spread serves as an effective indicator of forecast uncertainty. Ideally, this spread should align with the position error of the ensemble mean \cite{Yamaguchi2009}. Firstly, the accuracy of ensemble mean track predictions is evaluated by calculating the RMSE of distances between the ensemble mean and the best TC track values, denoted as $\textrm{ERROR}_{TC}$. Subsequently, the effectiveness of forecast uncertainty on track predictions, using the ensemble spread, is determined by the RMSE of the distances between individual ensemble members and the ensemble mean \cite{whitaker1998relationship}, represented as $\textrm{SPRD}_{TC}$.
The equations are presented below:
\begin{equation} 
\label{error_equation}
     \textrm{ERROR}_{TC} = \sqrt{\bar{p}_{forecast} - p_{o}}
\end{equation}
     % \textrm{ERROR}_{TC} = \vert \bar{p}_{forecast} - p_{o} \vert

\begin{equation} 
\label{spread_TC_equation}
     \textrm{Spread}_{TC} = \sqrt{\frac{1}{N} \sum_{n=1}^{N} (\bar{p}_{forecast} - p_{forecast}(n))}
\end{equation}
where $N$ is the number of ensemble members, $\bar{p}_{forecast}$ is the ensemble mean, and $p_{forecast}(n)$ is the track forecast of member $n$. $p_{o}$ is the best track value from IBTrACS. The sum of $\textrm{ERROR}_{TC}$ and $\textrm{Spread}_{TC}$ from the forecast initialization time to the prediction time, with time intervals of 6 hours, is defined as the accumulated ensemble mean position error and ensemble spread.
\begin{equation} 
\label{AccERROR_equation}
     \textrm{AccERROR}_{TC} =  \sum_{t \in T} \textrm{ERROR}_{TC}(t)
\end{equation}

\begin{equation} 
\label{AccSpread_equation}
     \textrm{AccSpread}_{TC} = \sum_{t \in T} \textrm{Spread}_{TC}(t)
\end{equation}
where t ranges from 6 to 120 hours. In an ideal EPS, the $\textrm{AccSpread}_{TC}$ and the $\textrm{AccERROR}_{TC}$ should be equal. If $\textrm{AccSpread}_{TC}$ exceeds $\textrm{AccERROR}_{TC}$, it indicates that the ensemble spread surpasses the mean error, suggesting an overestimation of forecast uncertainties. Conversely, if $\textrm{AccSpread}_{TC}$ is less than $\textrm{AccERROR}_{TC}$, it signifies that the ensemble spread is insufficient, leading to an underestimation of forecast uncertainties.

Moreover, TC position error can be decomposed into along-track ($\textrm{AT}_{TC}$) and cross-track ($\textrm{CT}_{TC}$) components, each reflecting different aspects of error orientation \cite{neumann1981models,goerss2000tropical,aberson2001ensemble}. The $\textrm{AT}_{TC}$ error, often referred to as timing error, quantifies discrepancies in the timing or speed of the TC's movement. The $\textrm{CT}_{TC}$ error, also known as directional error, measures biases in the direction perpendicular to the TC's trajectory. Consequently, significant $\textrm{AT}_{TC}$ errors can affect the timing of TC landfall, while substantial $\textrm{CT}_{TC}$ errors may influence the exact location or even occurrence of landfall \cite{Nicholas2021}. Specifically, an $\textrm{AT}_{TC}$ error is negative (positive) if the forecast TC position is behind (ahead) of the observed TC position along the TC track. In terms of $\textrm{CT}_{TC}$ error, a negative (positive) value indicates that the predicted position lies to the left (right) side of the observed track in the Northern Hemisphere, and this orientation reverses in the Southern Hemisphere. Therefore, $\textrm{AT}_{TC}$ errors provide insight into whether the forecasted TC motion is slower or faster than observed, while $\textrm{CT}_{TC}$ errors help determine whether the forecast inaccurately predicts the TC's recurvature timing.

\section*{Data Availability Statement}

We downloaded a subset of the ERA5 dataset from the official website of Copernicus Climate Data (CDS) at \url{https://cds.climate.copernicus.eu/}. The ECMWF ensemble mean and ensemble std are available at \url{https://apps.ecmwf.int/archive-catalogue/?type=em&class=od&stream=enfo&expver=1}. We downloaded the ECMWF ensemble data from the public cloud storage bucket for WeatherBench 2 at the link \url{https://console.cloud.google.com/storage/browser/weatherbench2/datasets/ifs_ens}.

%NOAA Interpolated Outgoing Longwave Radiation (OLR) were downloaded from \url{https://psl.noaa.gov/data/gridded/data.olrcdr.interp.html}

%去掉谷歌网盘链接
\section*{Code Availability Statement}
The source code employed for training and running FuXi-ENS models in this research is accessible within a specific Google Drive folder (\url{https://drive.google.com/drive/folders/1z47CRQdKFZaOjtKQWSNZobC1_RePUVIK?usp=sharing}) \cite{code2024}.

The xskillscore Python package can be accessed at \url{https://github.com/xarray-contrib/xskillscore/}.

\section*{Acknowledgements}
This work was supported by the Postdoctoral Fellowship Program of CPSF under Grant Number GZB20240154 and National Key R\&D Program of China under Grant 2021YFA0718000, National Natural Science Foundation of China under Grant 42175052. We express our sincere appreciation to the researchers at ECMWF and Google for their invaluable efforts in collecting, archiving, disseminating, and maintaining the ERA5 reanalysis dataset and ECMWF ensemble.

The computations in this research were performed using the CFFF platform of Fudan University.

%This research received support from the National Key R\&D Program of China under Grant 2021YFA0718000 and National Natural Science Foundation of China, Grant 42175052. 

\section*{Competing interests}
The authors declare no competing interests.

%\section*{Authors' contributions}
%L.C. and H.L. designed the research. L.C. performed the model training and evaluation. L.C. and X.Z. wrote the manuscript.

\noindent


\bibliography{refs}% common bib file


\end{CJK*}

\clearpage

\appendix

\renewcommand\thefigure{\thesection.\arabic{figure}}    

\setcounter{figure}{0}    

\end{document}